# A HMAX with LLC for Visual Recognition

Kean Hong Lau, Yong Haur Tay, Fook Loong Lo

Centre for Computing and Intelligent System

Universiti Tunku Abdul Rahman, Kuala Lumpur, Malaysia
{laukh,tayyh,lofl}@utar.edu.my

**Abstract.** Today's high performance deep artificial neural networks (ANNs) rely heavily on parameter optimization, which is sequential in nature and even with a powerful GPU, would have taken weeks to train them up for solving challenging tasks [22]. HMAX [17] has demonstrated that a simple high performing network could be obtained without heavy optimization. In this paper, we had improved on the existing best HMAX neural network [12] in terms of structural simplicity and performance. Our design replaces the L1 minimization sparse coding (SC) with a locality-constrained linear coding (LLC) [20] which has a lower computational demand. We also put the simple orientation filter bank back into the front layer of the network replacing PCA. Our system's performance has improved over the existing architecture and reached 79.0% on the challenging Caltech-101 [7] dataset, which is state-of-the-art for ANNs (without transfer learning). From our empirical data, the main contributors to our system's performance include an introduction of partial signal whitening, a spot detector, and a spatial pyramid matching (SPM) [14] layer.

## 1    Introduction

Current deep ANNs rely heavily on weights optimization that requires a huge amount of training data in order to generalize well. The Caltech-101 dataset [7], with a limited training sample size of maximum 30 images per category, presents a big challenge to ANNs. ANN performances on this chart has been limited in the early years. This is also the earliest dataset with a practically large number of object classes (102 of them) for large scale object recognition testing. Other datasets that have more classes include Caltech-256 [10] and ImageNet [4], but they were introduced much later and thus have less benchmarks for comparison. In this paper, we will focus on Caltech-101 for object classification due to these challenging attributes, as well as being one of the most popular and heavily benchmarked datasets, making it suitable for extensive method-to-method comparison.

With the recent arrival of a huge dataset called ImageNet, having millions of images and thousands of classes, one could pre-train a neural network with it, and then use the learned features to retrain the back layers of the network on a (different) target dataset. This is now known as the transfer learning approach, which recently gave state-of-the-art performances in many datasets [5, 11, 22]. However, we will ignore transfer-learning methods in our comparison because transfer learning may give an unfair performance advantage over the existing (single dataset) benchmarking protocol. This is because the two datasets involved in transfer learning may have significant overlaps For example, the images from both Caltech-101 and ImageNet came from the same source, namely the Internet. It is not clear how much of the images in Caltech-101 could be found in the huge ImageNet dataset. To avoid this domain overlapping complication, we will focus only on non-transfer learning methods.

Our research focuses on ANN as we believe it has aspects that are yet to be explored and improved. Here, we would like to advance the ANN technology in terms of the network architecture and recognition accuracy.

Disregarding transfer learning, ANN methods that scored high on Caltech-101 are designed by Goh *et al.* [9] at 78.9%, Sohn *et al.* [18] at 77.8%, and Hu *et al.* [12] at 76.1%. The system in [9] involves an unsupervised learning phase with the restricted Boltzmann machine (RBM) followed by a supervised optimization phase on the RBM features; the system in [18] involves a series of initialization procedures using clustering and Gaussian mixed model (GMM) followed by a contrastive divergence iteration; and finally, system [12] performed PCA learning for the filters of the front convolutional layer, and requires a sparse coding (SC) optimization for nodes' activation towards the back layer. All three methods above involve some kind of heavy iterations. We identify that heavy iterations is perhaps one of the drawbacks of current ANNs compared to some of the more efficient AI methods available [2, 15, 20]. In this work, we would like to improve on the computational efficiency of a neural network, as well as its accuracy. Of the three ANN approaches above, we identified that HMAX has the highest potential of being efficient and iteration-free. This is because the PCA involved in the HMAX is a relatively simpler process and should be easier to be simplified compared to the RBM theory. As for the SC module, there exist approximations or equivalences that are more efficient, such as the *k*-th nearest neighbors (*k*-NN) [2], the orthogonal matching pursuit (OMP) [3, 19] and the LLC [20]. Therefore, we have decided to pursue along the HMAX line of research.

Our study led us to a simple and yet powerful HMAX design. The final design involves replacing the PCA filters with orientation edge filters, including both greyscale and color space matching, and replacing SC with LLC. Our results showed that this combination could perform as well as the existing HMAX without the needs of PCA preprocessing, SC optimization and dictionary fine-tuning, all of which demand sequential computation and long iterations of convergence. Our experiments showed that the final system gave state-of-the-art performance on Caltech-101.

## 2    Related Work

We present a survey of the HMAX architectures in this section. Riesenhuber and Poggio [17] introduced the HMAX neural network architecture (see Figure 1) for solving object recognition. This network uses the Gabor filters as the first level linear filters with a circular mask. There are six levels, including the first layer representing the input image. It consists of alternating S (simple) and C (complex) layers, followed by a final support vector machine (SVM) classifier as the output layer. The first S layer S1 performs a convolution of the input image with Gabor filters of 16 various scales per orientation. There are four orientations, and both positive and negative filters may be included for opposite contrast detection. The next C1 layer does an 8×8 neighborhood max pooling over the S1 layer and over every pair of adjacent scales. The next S2 layer contains 256 2×2 C1 cell neighborhood patterns. A set of *p* patches (aka. templates) of various sizes randomly sampled over the C1 layer was used. Each S2 node stores a template of the training sample. The Gaussian radial basis function (GRBF) was used as the activation function. A final C2 global max pooling was computed over the S2 output giving *p* features per orientation. The final view-tuned unit (VTU) layer learns from the C2 activation patterns of all training samples and uses that to predict the class of an unknown test image. The parameters used in S1 and C1 are designed based on biological data. This architecture achieved 42% on Caltech-101.

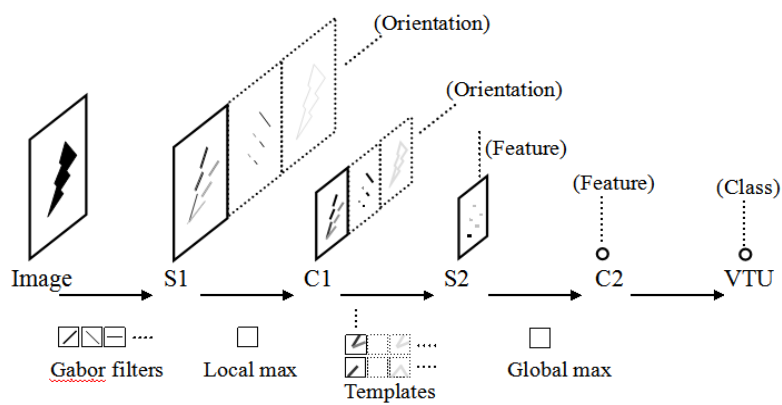

**Fig. 1.** The basic HMAX architecture. Input comes from the left and propagates towards the right. With convolution filters or templates, the S layers space would expand in proportion to the number of filters or templates used. The VTU is usually implemented with an SVM classifier.

Mutch and Lowe [16] extended the original HMAX architecture with sparse features. Sparsity is introduced by three main computations. The first computation involves lateral inhibition to suppress weak activity at all S and C layers, whereby nodes with activation value less than 50% of the maximum activation orientation of a particular location is suppressed to zero. In the second computation, the Gaussian match function is computed only for the dominant (maximum) orientations of a template at a particular position. The S2 templates are randomly sampled from images of the training set. Lastly, low weight templates leading to the VTU are discarded, and they noticed that small receptive fields survive better than larger ones. Their Caltech-101 performance was 56% with an SVM classifier.

Hu *et al.* [12] made an even bigger improvement over the original HMAX model by replacing the GRBF encoding with SC activation and dictionary learning. The first level of convolutional filters (Gabor filters) are replaced by PCA filters trained from the Kyoto dataset (a kind of transfer learning, but limited to the lowest level), and the C2 global max pooling with a 3-level SPM [12] subregions having resolutions of {1×1, 2×2, 4×4}. Performance as high as 76.1% was reported with a network aggregation of two HMAXs, one with two S-C stacks and another one with three S-C stacks. Single HMAX performance was only 73.7%.

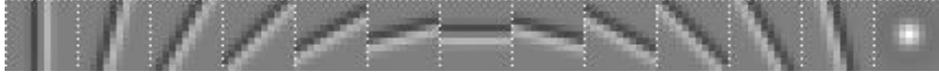

**Fig. 2.** Visualization of the S1 filters consists of 12 edge orientations and a spot pattern.

## 3   Approach

We experimented with the HMAX architecture proposed by Hu *et al.* [12]. We replaced the SC module with LLC. LLC has achieved many good results in vision [8, 20]. It is a type of encoding that leads to sparse outputs. Sparse coding has been observed in the brain [12], and such property may be desirable in an information processing system. Our LLC implementation is based on the approximated LLC algorithm as suggested in [20] but with some modifications and simplifications. Our algorithm begins by collecting all *k* nearest-neighbor templates to an incoming patch *x* to form the local bases of a compact dictionary **D** (with *k* entries) and solve for the activation codes *c* (an array of *k* elements) by,

$$\min_c \|x - \mathbf{D}c\|^2. \quad (1)$$

Instead of distance measure, we use the dot-product to determine template closeness in the above *k*-NN search. We also remove the requirement that $\mathbf{1}^T c = 1$, which reduces the above minimization to a simple and fast least-squares (LS) solution. This formulation is much more efficient than the L1 minimization convergence search. In particular, when *k* is small (typically 15 as will be seen later), the LS minimization involves manipulating a small *k*×*k* matrix.

All train and test images are reduced to a chosen maximum number of pixels per side, while maintaining the original image aspect ratio. The resultant images are contrast-stretched to maximize the dynamic range of color saturation while maintaining the original RGB ratio (hue), before feeding them to the HMAX network. In the first layer of the network, the RGB channels are converted into opponent color channels (i.e. intensity, R-G and Y-B) [23], convolved with an S1 edge filter respectively, and averaged up their respective output magnitudes as the output value. In a typical setting, we use up to 12 11×11 S1 orientation edge filters evenly spread over 180° (see Figure 2). The edge filters are constructed using the first derivative of Gaussian, which we found are as effective as Gabor filters, but without the strong edge ripple effects (i.e. periodic repetitive parallel lines) usually observed in the output of a Gabor filter convolution process, and we think that this lack of frequency interference is desirable. We also included one optional spot pattern for spot detection as spots have also been reported in the learnt kernels of ANNs [22], among the edge patterns. All filters (and templates) are whitened and normalized to unit length. The incoming input patch patterns are, however, partially whitened and normalized before convolution. In our experiments, we discovered that a 98% whitening formula worked best, together with a semi-saturation constant $\beta$ (similar to the constant $\sigma^2$ used in equation (2) of [23]). The formula for this transformation is given in (2) below.

$$y = (x - \alpha\mu)/(\sigma + \beta) \quad (2)$$

where *y* is the output response of partial whitening followed by unit length normalization with saturation $\beta$ for an incoming signal *x* over a small (convolution) window patch of a layer. $\alpha$ refers to the fraction of a full whitening process, (typically set to 0.98). $\mu$ and $\sigma$ are the mean and standard deviation of signal *x*. The value of $\beta$ is not so critical in practice, we have tried values from 0 to 20 and cannot find a conclusive optimum. In our default setting, we chose $\beta = 3$.

The output of S1 is then pooled using a local max operator of size 12×12 and sub-sampled by a ratio of 6:1 at the C1 layer.

We randomly sample, from C1, thousands of 4×4 S2 templates generated by the training image set. All templates are whitened and normalized, whereas the incoming signal from C1 undergoes a similar pre-processing step as in (2) before being operated upon at S2 with LLC *k*-NN convolution and LS minimization. For LS, we use a ridge regression penalty coefficient of 0.25.

Two SPM structures are generated at C2 using max pooling. The first SPM has two levels with 1×1 and 3×3 sub-regions pooling from S2; while the second SPM has three levels with 1×1, 2×2, and 4×4 respective sub-regions, each further splits into a positive and a negative polarity output bins. So, each S2 feature (template) gives rise to 52 C2 feature points.

All the pooled feature points from all templates at C2 are inputted into a logistic regression classifier first for training and later for prediction. We use the LIBLINEAR [6] executable module, with the cost parameter set to 0.1, which we found to normally give higher scores than other settings.

The typical time for one single-threaded run through our architecture (includes initialization, training and testing) on Caltech-101 is about 14 hours with 30 training samples and up to 50 testing images per class, using 1000 templates. Our computers are multithreading dual-core i5 processors clocking at 3.0 GHz and all timings were measured on single thread (without any GPU processing).

## 5   Comparison with Other ANNs

Some differences between our HMAX and the previous HMAX versions are summarized below:

- Uses the first derivative of Gaussian as the lowest level convolution filter, vs Gabor or PCA filters.
- Partial whitening and unit normalization of all input signals to the S layers, and fully whitening and normalization of all convolution filters and templates.
- Opponent color space processing at S1 layer, vs greyscale processing.
- Uses LLC as the encoding scheme compared to GRBF or L1 minimization encoding at S2.
- Deep SPM of 4 levels with polarity distinctions, compared to a single level or 3-level SPM in the previous designs.

Comparing with convolutional network and RBM, the main distinctions are:

- Pre-designed edge filters and direct incoming pattern sampling for templates, vs random initialization and optimization iterations (typically of the gradient descend nature).
- No bias signals used in the convolution process, and no sigmoid function used at node outputs.

- Makes use of an over-complete set with thousands of features at the second S-C stack (which is typically the second level of convolution) vs typically hundreds filters or less for the same level of a convolutional network.

## 6 Experimental Results

### 6.1 Parameter Range Searching

Following the procedure in [7], we select 30 samples randomly from each training class, and test with up to 50 samples per class. All figures below are averaged over at least 5 trials. All experiments are carried out with the default settings, unless otherwise stated, of 1000 S2 templates, 240 image resolution, LLC pooling size of 20, and using a deep SPM with opponent color space edge detection.

Figure 3(i) shows that performance increases with image resolution but slows down towards the maximum level of 300 pixels per side (most Caltech-101 images have a maximum resolution of 300×300). The reasons for the poor performance at low resolutions may be due to the loss of image fineness and details for matching.

In Figure 3(ii), the value of neighborhood $k$ (in $k$-NN of LLC) covers a wide range of values from 5 to 30 without much performance deviation, such phenomenon was also observed in [20] who reported that performance peaked at 5, with bigger figures came very close, except 2 which gave an apparent score drop. However, our best score occurs at 15, which could be due to our features not undergoing any clustering reduction procedure, thus needing a bigger pool size for a similar performance.

In Figure 3(iii), the performance increases albeit slowly with respect to the number of templates used. There seems no apparent break to the climb, indicating a possibility of ever improving the performance with more templates. However, an over-complete set of features would demand memory and computation resources that could be prohibitive in practice, and may not worth paying for the tiny increment gained. This plot also demonstrates that our system could still reliably perform well with a small set of templates without significant accuracy drop (with less than 2% reduction going from 2500 templates down to 500).

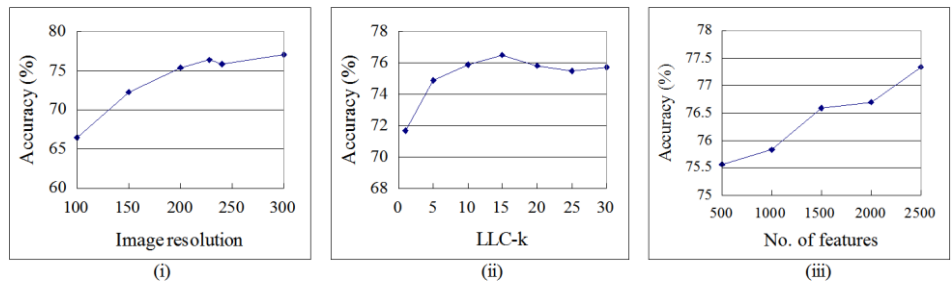

**Fig. 3.** (i) The Caltech-101 classification accuracy against image resolution measured by the maximum side dimension of an image. (ii) Classification accuracy vs LLC's neighborhood size $k$. (iii) Classification accuracy vs number of templates used in the S2 layer.

|                | Accuracy (%) |
|----------------|--------------|
| Full model     | 75.83        |
| Greyscale only | 75.70        |
| Global max     | 60.22        |
| 3-level SPM    | 75.76        |
| No whitening   | 73.48        |
| Full whitening | 72.68        |

**Table 1.** Determining the contributions of color, deep SPM, and whitening to system accuracy. The "Full model" includes all 3 components above. Each subsequent row has one of the 3 components removed or modified.

### 6.2 Performance Contributing Factors

We ran a series of experiments to check on the contributions of the components of the network to the overall performance, and the results are shown in Table 1. We want to see how much color contributes to the performance. Comparing scores between "Full Model" and "Greyscale only", we see that color contributes roughly to +0.13% compared to using just greyscale alone. This shows that greyscale is already quite informative for edge detection, and color only improves incrementally.

Comparing "Global max" (which is a special case of SPM with only one level and a single region) with "3-level SPM", we can see a big drop of -15.61% if the full standard 3-level SPM is not used. Another observation is that deep SPM helps only a little by +0.07%. Thus, we can say that the standard 3-level SPM commonly implemented in the vision field is very effective in general.

Comparing the "Full model" with the last two row entries confirms that our 98% whitening process is respectively +2.35% and +3.15% better than either no whitening or full whitening. We believe partial whitening helps in conditioning the incoming signal to a narrower dynamic range around the origin for subsequent stage processing. This limited range helps to improve signal sensitivity for the next level computation. However, fully removing the mean of the signal might have completely removed all useful information. Since HMAX does not include a bias term in its layer (unlike convolutional network where the bias weights could be learnt), such an additional control on signal dc level helps in this architecture.

### 6.3 Benchmark Comparison

In our final series of experiments, we combine the optimum values of image resolution (300×300) and the $k$ parameter (15) in a final system to check for any improvement. We included three large feature sizes (2000, 3000 and 4000) for testing, as suggested in Figure 3(iii) where bigger size gives higher score. We also experimented with edge filters of various number of orientation filters (4, 8, and 12) with or without a spot detector. A comparison with other published scores is shown in Table 2 where we reported the top three configurations found in our experiments. HMAX(LLC) architecture I consists of 8 orientation filters with a spot detector and 3000 templates; architecture II consists of 12 orientation filters without the spot detector and 2000 templates; and architecture III consists of 12 orientation filters without the spot detector and 4000 templates.

Our score of 79.0% using 3000 templates (architecture I) is the best for a HMAX on Caltech-101 and has improved 2.9% over the previous best HMAX architecture. This score is also the state-of-the-art for ANN on Caltech-101 under the category of non-transfer learning. However, this is still 4.3% below the current best AI method [21], but we hope future improvements would narrow down the gap or exceed it. The large feature size of the current system could also be a hindrance to practicality, so future works for improvement would include template pool size reduction. We also foresee a need to test our system on bigger datasets like Caltech-256 and ImageNet.

Our system is relatively simple compared to other ANNs. This simplicity translates to a quicker initialization and training phase. Our system could complete network training (time was mainly spent on logistic regression training) from the start in about 14 hours (for both architectures I and II) without a GPU running on single thread.

| Methods | Accuracy (%) |
|---|---|
| Transfer Learning (from ImageNet): | |
| He *et al.* [11] | 91.4 |
| Donahue *et al.* [5] | 86.9 |
| Zeiler and Fergus [22] | 86.5 |
| Non-Transfer Learning: | |
| Yang et al. [21] | 84.3 |
| Lim and Tay [15] | 83.5 |
| Feng *et al.* [8] | 82.6 |
| Bo *et al.* [1] | 82.5 |
| ANN (Non-Transfer Learning): | |
| HMAX(LLC) architecture I | **79.0 ± 0.7** |
| Goh *et al.* [9] | 78.9 |
| Sohn *et al.* [18] | 77.8 |
| HMAX(LLC) architecture II | 77.8 ± 0.7 |
| HMAX(LLC) architecture III | 77.5 ± 0.9 |
| Hu *et al.* [12] | 76.1 |
| Coates and Ng [3] | 72.6 |
| Jarret *et al.* [13] | 65.5 |

**Table 2.** Benchmark against other methods on Caltech-101. Our figures for HMAX(LLC) are averaged over at least 10 runs.

## 7    Discussion

Our approach presents an alternative design to the neural network architecture that is much simpler and structured. ANN has always been portrayed as a black box full of somewhat random weights, which could only be handled by a local gradient descend process working towards minimizing a certain error cost function and which in itself provides very little insights about the functions of the individual layers of an ANN and its global network structure. Recent developments go even more abstract with the highly iterative unsupervised RBM learning. Our HMAX architecture, on the other hand, is laying down flat and not hidden behind some heavy mathematical abstraction. It is hinting to us that the bottom stack is extracting small local shape features like edges and spots, while deeper layers are computing intermediate (4×4 S2 templates) pattern matches (compare this to the pieces of a jigsaw puzzle), while the last stage is combining all these matches to reach a final decision, and it also tells us that locations of the features are important (supported by the 4×4 sub-regions of an SPM). Therefore, apart from the computational speed-up advantage, we also hope that this structural simplicity that we have designed and shown to perform so well would contribute to the study and understanding of the significance of each layer and weight of an ANN as a computing machine. Such understanding could help to better initialize the network weights (replacing RBM iterations) and to speed up the convergence of ANNs, or to remove completely the need for network weights fine-tuning all together. However, one would point out that the logistic regression is also an iterative optimization process. True, but this stage is relatively shallow and computationally light in practice. And if so desired, this layer could be replaced by a linear LS approximation which has a closed form solution, with perhaps a small sacrifice on accuracy performance.

## 8  Conclusion

We had improved on the HMAX architecture in performance as well as computational efficiency. In our design, the first level filter bank has a pre-designed weight pattern created using the first derivative of Gaussian, there is no training required. In the second level filter bank, our system randomly samples small patches of the incoming signal patterns and treat them as the filters (templates). There are no optimization steps required for further filter weights fine-tuning, and thus making both processes above computationally efficient. Computation is further reduced with a simplified LLC approximation that involves a $k$-NN and a LS minimization.

We achieved very good performance with an over-complete set of thousands of templates, and yet the system could still perform robustly well even when the number of templates has been reduced by an order. Our system's performance has achieved state-of-the-art in ANNs, without using transfer learning, on Caltech-101. Apart from more features, the good performance is also due to the SPM, a better signal conditioning with signal partial whitening, and the inclusion of a spot detector at S1. Future research should focus on feature size reduction, as well as testing the system on other datasets.